\pdfoutput=1
\documentclass[runningheads]{llncs}

\usepackage[noadjust]{cite}
\usepackage{amsmath,amssymb,amsfonts,bm}
\usepackage{algorithmic}
\usepackage{textcomp}
\usepackage{xcolor}
\usepackage{microtype}
\usepackage{multirow,array}
\usepackage{graphicx}
\usepackage[font=tiny]{subcaption}
\usepackage{subcaption}
\usepackage{times}
\usepackage{latexsym}
\usepackage[export]{adjustbox}

\newcommand{\tabspace}{\,\,\,\,\,\,\,\,}

\newcommand{\mycomment}[1]{}
\begin{document}
\title{Adversarial Capsule Networks for Romanian Satire Detection and Sentiment Analysis}
%

\author{Sebastian-Vasile Echim\inst{1} \and
Răzvan-Alexandru Smădu\inst{1} \and
Andrei-Marius Avram\inst{1} \and
Dumitru-Clementin Cercel\inst{1} \and
Florin Pop\inst{1,2}
}
\institute{Faculty of Automatic Control and Computers, University Politehnica of Bucharest \and
National Institute for Research and Development in Informatics - ICI Bucharest, Romania
\email{sebastian.echim@stud.aero.upb.ro
\{razvan.smadu,andrei\_marius.avram\}@stud.acs.upb.ro
\{dumitru.cercel,florin.pop\}@upb.ro}
}

\authorrunning{S.-V. Echim et al.}
%

\maketitle
\begin{abstract}

Satire detection and sentiment analysis are intensively explored natural language processing (NLP) tasks that study the identification of the satirical tone from texts and extracting sentiments in relationship with their targets. In languages with fewer research resources, an alternative is to produce artificial examples based on character-level adversarial processes to overcome dataset size limitations. Such samples are proven to act as a regularization method, thus improving the robustness of models. In this work, we improve the well-known NLP models (i.e., Convolutional Neural Networks, Long Short-Term Memory (LSTM), Bidirectional LSTM, Gated Recurrent Units (GRUs), and Bidirectional GRUs) with adversarial training and capsule networks. The fine-tuned models are used for satire detection and sentiment analysis tasks in the Romanian language. The proposed framework outperforms the existing methods for the two tasks, achieving up to 99.08\% accuracy, thus confirming the improvements added by the capsule layers and the adversarial training in NLP approaches.

\end{abstract}

\section{Introduction}

Satirical news is a type of entertainment that employ satire to criticize and ridicule, in a humorous way, the key figures from society, socio-political points, or notable events~\cite{yang-etal-2017-satirical,rogoz-etal-2021-saroco}. Although it does not aim to misinform, it mimics the style of regular news. Therefore, it has a sizeable deceptive potential, driven by the current increase in social media consumption and the higher rates of distrust in official news streams \cite{mchardy-etal-2019-adversarial}. 

Furthermore, sentiment analysis is regarded as a successful task in determining the opinions and feelings of people, especially in online shops where customer feedback analysis can lead to better customer service~\cite{xuetal2019sentiment}. Limited resources in languages such as Romanian make it challenging to develop large-scale machine learning systems since the largest datasets present up to tens of thousands of examples~\cite{rogoz-etal-2021-saroco}. Therefore, various techniques should be proposed and investigated to address these challenges on such datasets. 

Adversarial training is an effective defense strategy to increase the robustness and generalization of the models intrinsically. Introduced by Szegedy et al.~\cite{szegedy2013intriguing} and analyzed by Goodfellow et al.~\cite{goodfellow2015explaining}, adversarial examples are augmented data points generated by applying a small perturbation to the input samples. It was initially employed in computer vision, where input images were altered with a small perturbation \cite{goodfellow2015explaining,xiao2018generating,MadryMSTV18}. More recently, adversarial training gained popularity in NLP. The text input is a discrete signal; therefore, the perturbation is applied to the word embeddings in a continuous space \cite{miyatoadversarial}. The application of adversarial training in our experiments is motivated by the potential to improve the robustness and generalization of models with limited training resources. 

This paper aims to introduce robust high-performing networks employing adversarial training and capsule layers~\cite{sabour-etal-2017} for satire detection in a Romanian corpus of news articles~\cite{rogoz-etal-2021-saroco} and sentiment analysis for a Romanian dataset~\cite{tache-etal-2021-loresda}.  
Our experiments include training models suitable for NLP tasks as follows: Convolutional Neural Networks (CNNs)~\cite{zhang2015sensitivity}, Gated Recurrent Units (GRUs)~\cite{cho2014properties}, Bidirectional GRUs (BiGRUs), CNN-BiGRU, Long Short-Term Memory (LSTM)~\cite{hochreiter1997long}, Bidirectional LSTM (BiLSTM), and CNN-BiLSTM.
Starting from Zhao et al.~\cite{zhao2019capsule}, we compare the networks against their adversarial capsule flavors. Next, the best-performing network is subjected to an in-depth analysis concerning the impact on the performance of the capsule model and the training with adversarial examples. Thus, we test the effect of capsule hyperparameters varying the number of primary and condensed capsules~\cite{zhao2019capsule}. Also, we assess the performance of our model employing Romanian GPT-2 (RoGPT-2)~\cite{niculescu2021rogpt2} for data augmentation up to 10,000 text continuation examples. Finally, we discuss several misclassified test inputs for the sentiment analysis task.

The main contributions in this work are as follows:

(i) we thoroughly experiment with various configurations to assess the performances of the investigated approaches, namely adversarial augmentations and capsule layers;
(ii) we show that the best-performing model uses BiGRU with capsule networks, while the most improvements were seen when incorporating RoGPT-2-based augmentations;
(iii) we investigate the effects of analyzed components through t-SNE plots~\cite{van2008visualizing} and ablation studies;
and
(iv) we achieve state-of-the-art results on the two Romanian datasets.

\section{Related Work}

\subsection{Capsule Networks in NLP}

Firstly presented by Sabour et al.~\cite{sabour-etal-2017}, the capsule neural networks are machine learning systems that model hierarchical relationships regarding object properties (such as pose, size, or texture) in an attempt to resemble the biological structure of neurons. Among other limitations, capsule networks are addressing the max pooling problem of the CNNs, which allows for translation invariance, making them vulnerable to adversarial attacks \cite{patrick2022capsule}. While it has been demonstrated that capsule networks are successful in image classification~\cite{sabour-etal-2017}, there is also a general preference for exploring their potential in NLP tasks, especially in text classification. Several works \cite{kim-etal-2018-capsules, zhao-etal-2018-capsules} took the lead in this topic, showing that using different approaches, such as static and dynamic routing, the capsule models provided competitive results on popular benchmarks.

Several studies were performed in topic classification and sentiment analysis using capsule networks. Srivastava et al.~\cite{srivastava-etal-2018-identifying} addressed the identification of aggression and other activities, such as hate speech and trolling, using a model based on the dynamic routing algorithm~\cite{zhao-etal-2018-capsules} involving LSTM as a feature extractor, two capsule layers (namely, a primary capsule layer and a convolutional capsule layer), and finally, the focal loss~\cite{lin-etal-2017-focal} to handle the class imbalance. The resulting model outperformed several robust baseline algorithms in terms of accuracy; however, a more complex data preprocessing was expected to improve the results further.

For the sentiment analysis task, Zhang et al.~\cite{zhang-etal-2018-capsule-semantic-rules} proposed CapsuleDAR, a capsule model successfully combined with the domain adaptation technique via correlation alignment~\cite{sun2017correlation} and semantic rules. The model architecture consisted of a base and a rule network. The base network employed a capsule network for sentiment prediction, consisting of several layers: embedding, convolutional, capsule, and classification. The rule network involved a rule capsule layer before the classification layer. Extensive experiments were conducted on review datasets from four product domains, which showed that the model achieved state-of-the-art results. Additionally, their ablation study showed that the accuracy decreased sharply when the capsule layers were removed.

Su et al.~\cite{su-etal-2020} tackled limitations of Bidirectional Encoder Representations from Transformers (BERT)~\cite{devlin2018bert} and XLNet~\cite{yang2019xlnet}, such as local context awareness constraints, by incorporating capsule networks. Their model considered an XLNet layer with 12 Transformer-XL blocks on top of which the capsule layer extracted space- and hierarchy-related features from the text sequence. Experiments illustrated that capsule layers provided improved results compared with  XLNet, BERT, and other classical feature-based approaches. 

Moreover, Saha et al.~\cite{saha-etal-2020} introduced a speech act classifier for microblog text posts based on capsule layers on top of BERT. The model took advantage of the joint optimization features of the BERT embeddings and the capsule layers to learn cumulative features related to speech acts. The proposed model outperformed the baseline models and showed the ability to understand subtle differences among tweets.

\subsection{Romanian NLP Tasks}

In recent years, several datasets have emerged aiming to improve the performance of the learning algorithms on Romanian NLP tasks. Apart from the two datasets used in this work, researchers have also introduced the Romanian Named Entity Corpus (RONEC)~\cite{dumitrescu2020introducing} for named entity recognition\footnote{A new version of RONEC is available at \url{https://github.com/dumitrescustefan/ronec}}, the Moldavian and Romanian Dialectal Corpus (MOROCO)~\cite{butnaru2019moroco} for dialect and topic classification, the Legal Named Entity Recognition corpus (LegalNERo)~\cite{pais2021named} for legal named entity recognition, and the Romanian Semantic Textual Similarity dataset (RoSTS)\footnote{\url{https://github.com/dumitrescustefan/RO-STS}} for finding the semantic similarity between two sentences.

Lately, the language model space for Romanian was also improved with the introduction of Romanian BERT (BERT-ro)~\cite{dumitrescu2020birth}, RoGPT-2, ALR-BERT~\cite{nicolae2022lite}, and DistilMulti-BERT~\cite{avram2021distilling}. In addition, all the results for these systems have been centralized in the Romanian Language Leaderboard (LiRo)~\cite{dumitrescu2021liro}, a leaderboard similar to the General Language Understanding Evaluation (GLUE) benchmark~\cite{wang2018glue} that tracks over ten Romanian NLP tasks.

\section{Datasets}
In this work, we rely on two of the most recent Romanian language text datasets: a corpus of news articles, henceforth called SaRoCo~\cite{rogoz-etal-2021-saroco}, and one composed of positive and negative reviews crawled from a Romanian website, henceforth called LaRoSeDa~\cite{tache-etal-2021-loresda}.

\subsection{Satirical News}
SaRoCo is one of the most comprehensive public corpora for satirical news detection, eclipsed only by an English corpus \cite{yang-etal-2017-satirical} with 185,029 news articles and a German one \cite{mchardy-etal-2019-adversarial} with 329,862 news articles. SaRoCo includes 55,608 samples, of which 27,628 are satirical and 27,980 are non-satirical (or regular). Each sample consists of a title, a body, and a label. On average, an entire news article has 515.24 tokens for the body and 24.97 tokens for the title. The average number of sentences and words per sentence are 17 and 305, respectively. The labeling process is automated, as the news source only publishes satirical or regular content.   

\subsection{Product Reviews}

LaRoSeDa is one of the largest corpora for sentiment analysis in the Romanian language. It was created based on the observation that the freely available Romanian language datasets were significantly reduced in size. This dataset totals 15,000 online store product reviews, either positive or negative, for which the ratings were also collected for labeling purposes. Thus, assuming that the ratings might reflect the polarity of the text, each review rated with one or two stars was considered negative. In contrast, the four or five-star labels were considered positive. The labeling process resulted in 7,500 positive reviews (235,474 words) and 7,500 negative reviews (304,813 words). The average number of sentences and words per review is 4 and 36, respectively.

\section{Methodology}

\begin{figure}[!ht]
  \centering
  \includegraphics[width=\textwidth]{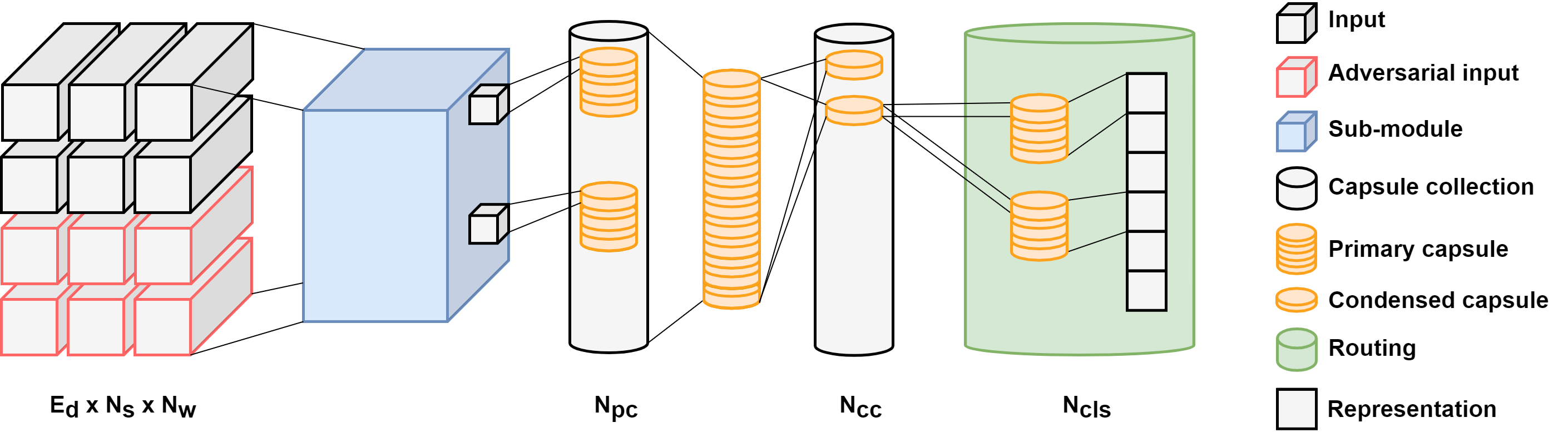}
  \caption{Our generic adversarial capsule architecture, where $E_d$ denotes the embedding size, $N_s$ is the number of sentences, $N_w$ is the number of words per sentence, $N_{pc}$ is the number of primary capsules, $N_{cc}$ is the number of condensed capsules, and $N_{cls}$ is the number of classes to which the routing algorithm will converge.}
   \label{fig:caps-archi}
\end{figure}

The generic adversarial capsule network we employ is presented in Figure~\ref{fig:caps-archi}. 
It consists of a sub-module that can represent any widely-used NLP model, followed by capsule layers. Concretely, we use primary capsules and capsule flattening layers to facilitate the projection into condensed capsules passed as input for a routing mechanism to obtain the class probabilities. To increase robustness, we feed regular and adversarial samples into the model. In what follows, we detail the employed components.

\textbf{Word Embeddings.}
Each word is associated with a fixed-length numerical vector, allowing us to express semantic and syntactic relations, such as context, synonymy, and antonymy. Depending on the model, the embedding representation has various sizes.

To use a continuous representation of the input data, we employ two different types of embeddings: BERT- and non-BERT-based.
On the RoBERT model~\cite{masala2020robert}, we rely on embeddings delivered by the model with a dimension $E_d=768$, whereas, for the non-BERT models, we abide by Onose et al.~\cite{onose-etal-2019-sc} in terms of distributed word representations and choose Contemporary Romanian Language (CoRoLa)~\cite{mititelu2018reference} with an embedding dimension $E_d=300$, Nordic Language Processing Laboratory (NLPL)~\cite{kutuzov2021large}, having the size $E_d=100$, and Common Crawl (CC)~\cite{graves2005phoneme} with $E_d=300$.

\textbf{Adversarial Examples.}
To increase the robustness of our networks, we create adversarial examples by replacing characters in words. Using the letters of the Romanian alphabet, we randomly substitute one character per word, depending on the sentence size: one replacement for less than five words per sentence, two replacements for 5 to 20 words per sentence, and three replacements for more than 20 words per sentence.

\textbf{Primary Capsule Layer.} 
This layer transforms the feature maps obtained by passing the input through the sub-module into groups of neurons to represent each element in the current layer, enabling the ability to preserve more information. By using $1 \times 1 $ filters, we determine the capsule $\bm{p}_i$ from the projection $p_{ij}$ of the feature maps~\cite{zhao2019capsule}:

\begin{equation}
\bm{p}_i = squash(p_{i1} \oplus p_{i2} \oplus \cdot\cdot\cdot \oplus p_{id}) \in \mathbb{R}^d \label{eq:3.4}
\end{equation} 
where $d$ is the primary capsule dimension, $\oplus$ is the concatenation operator, and $squash(\cdot)$ adds non-linearity in the model:

\begin{equation}
squash(\bm{x}) = \frac{ \|\bm{x}\|^2 }{1 + \|\bm{x}\|^2 }\frac{\bm{x}}{\|\bm{x}\|} \label{eq:3.5}
\end{equation}

\textbf{Compression Layer.}
Because it requires extensive computational resources in the routing process (i.e., the fully connected part of the capsule framework), we need to reduce the number of primary capsules. We follow the approach proposed by Zhao et al.~\cite{zhao2019capsule}, which uses capsule compression to determine the input of the routing layer $\bm{u}_j$. Each condensed capsule $\hat{\bm{u}}_j$ represents a weighted sum over all the primary capsules:

\begin{equation}
\hat{\bm{u}}_j = \sum_{i} b_i \bm{p}_i \in \mathbb{R}^d \label{eq:3.6}
\end{equation}

\textbf{Routing Layer.} 
It conveys the transition layer between the condensed capsules to the representation layer. It is denoted by a routing method to overcome the loss of information determined by a usual pooling method. In our capsule framework, we choose Dynamic Routing with three iterations \cite{sabour-etal-2017}.

\textbf{Representation Layer.} 
In the binary classification tasks, the last slice of our generic architecture is represented by the probability of a text input being satirical or regular for SaRoCo and positive or negative sentiment for LaRoSeDa.

\section{Experimental Setup}

\subsection{Model Parameters}
Firstly, we use CoRoLa, CC featuring 300-dimensional, and NLPL with 100-dimensional state space vectors for reconstruction at the embeddings level. We choose n-gram kernels with three sizes (i.e., 3, 4, and 5) and 300 filters each for the CNN sub-module. Also, for the Capsule layers, we use $N_{pc}=8$ primary capsules and $N_{cc}=128$ condensed capsules, which we fully connect through Dynamic Routing and obtain $N_t$ lists with $N_{cls}$ elements. For each element in the list, the argument of the maximum value represents the predicted label, where ``1" is a satirical text or a positive review, whereas ``0" is a non-satirical text or a negative review. Secondly, for the GRU and LSTM sub-modules, we employ one layer and a hidden state dimension of 300 for both unidirectional and bidirectional versions. Finally, for the RoBERT model, we choose the base version of the Transformer with vector dimensions of 768, followed by a fully connected layer with the size of 64, $\tanh$ activation function, and a fully connected layer with $N_{cls}$ output neurons.

\subsection{Training Parameters}
The number of texts chosen from SaRoCo is $N_t=30,000$ (15,000 satirical and 15,000 non-satirical) with a maximum $N_s=5$ sentences per document and $N_w=60$ words per sentence. For LaRoSeDa, we use 6,810 positive and 6,810 negative reviews for training, with $N_s=3$ sentences per document and $N_w=60$ words per sentence. The optimizer is Adam~\cite{kingma2014adam}, and the loss function is binary cross-entropy. We set the learning rate to $5e-5$ with linear decay and train for 20 epochs. The batch size is 32, and the train/validation/test split is 70\%/20\%/10\%.

\section{Results}
This section presents the performance analysis of our models from quantitative and qualitative perspectives, as well as a comparison with previous works for the chosen datasets.

\textbf{Initial Results.}
Table~\ref{tab:training_results} shows our results on the SaRoCo and LaRoSeDa datasets. The experiments with varying embeddings other than RoBERT (i.e., CC, CoRoLa, and NLPL) show that NLPL determines better performance overall. This was unexpected because CoRoLa covers over one billion Romanian tokens, while CC and NLPL contain considerably fewer tokens. For the SaRoCo dataset, the best model on the CC embeddings uses the BiGRU sub-module, achieving a 95.80\% test accuracy. For the CoRoLa corpus, the GRU and BiGRU sub-modules perform equally, resulting in a 95.77\% test accuracy. Also, the best NLPL embedding model considers the BiGRU sub-module, scoring a 96.15\% test accuracy. On the LaRoSeDa dataset, we find the best model obtaining a 96.06\% test accuracy based on GRU with NLPL embeddings. Moreover, training on the RoBERT embeddings brings the highest performance when combined with the BiGRU sub-module, achieving a test accuracy of 98.32\% on SaRoCo and 98.60\% on LaRoSeDa.

The score differences between our results on the two datasets are less than 0.5\%. Therefore, a performance difference is expected due to the more considerable proportion of data for SaRoCo. Thus, there is no concrete insight into whether the satire detection task is more complex than the sentiment analysis one, especially in the binary classification setup. 
Still, since the training set size for LaRoSeDa is considerably smaller than that of the SaRoCo one, the slight performance difference shows polarization support on sentiment analysis.

\begin{figure}[!ht]
\centering
\begin{subfigure}[b]{0.1375\textwidth}
    \centering
    \caption{CNN}
    \includegraphics[width=\textwidth]{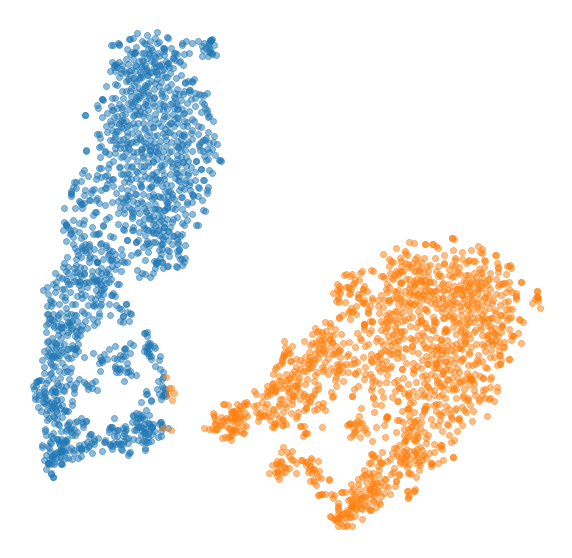}
    \label{fig:1-train-saroco}
\end{subfigure}
\begin{subfigure}[b]{0.1375\textwidth}
    \centering
    \caption{GRU}
    \includegraphics[width=\textwidth]{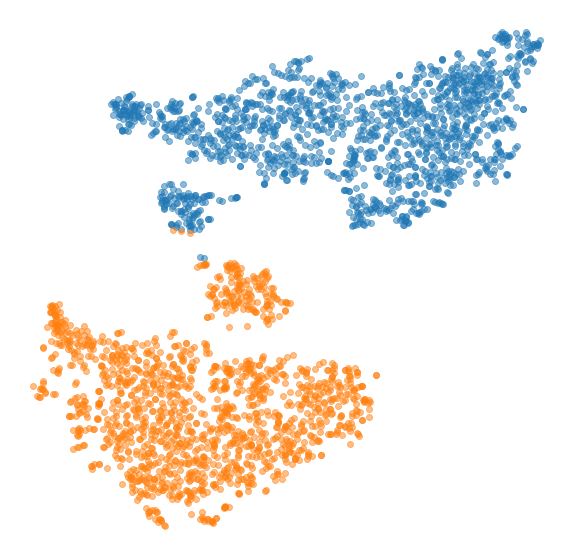}
    \label{fig:2-train-saroco}
\end{subfigure}
\begin{subfigure}[b]{0.1375\textwidth}
    \centering
    \caption{BiGRU}
    \includegraphics[width=\textwidth]{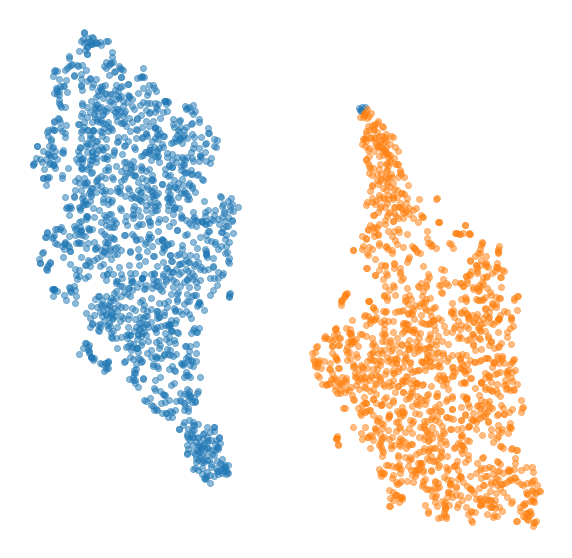}
    \label{fig:3-train-saroco}
\end{subfigure}
\begin{subfigure}[b]{0.1375\textwidth}
    \centering
    \caption{CNN-BiGRU}
    \includegraphics[width=\textwidth]{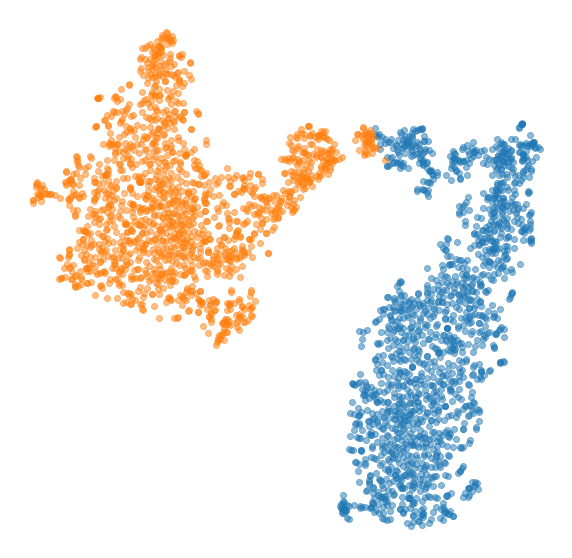}
    \label{fig:4-train-saroco}
\end{subfigure}
\begin{subfigure}[b]{0.1375\textwidth}
    \centering
    \caption{LSTM}
    \includegraphics[width=\textwidth]{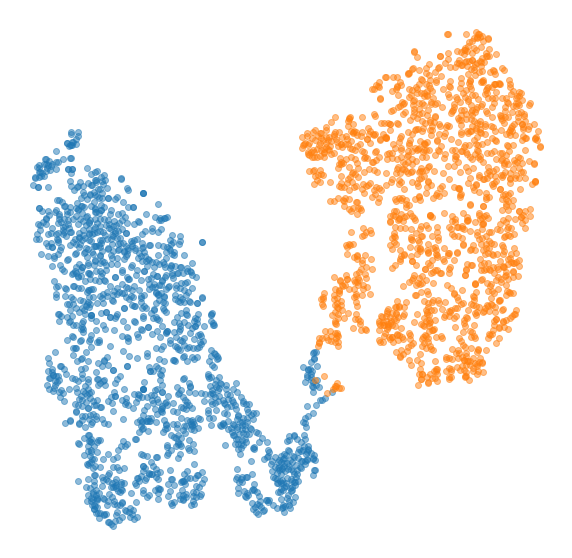}
    \label{fig:5-train-saroco}
\end{subfigure}
\begin{subfigure}[b]{0.1375\textwidth}
    \centering
    \caption{BiLSTM}
    \includegraphics[width=\textwidth]{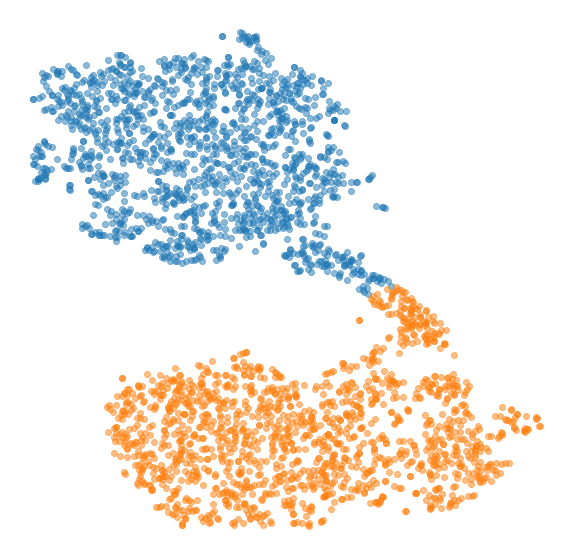}
    \label{fig:6-train-saroco}
\end{subfigure}
\begin{subfigure}[b]{0.1375\textwidth}
    \centering
    \caption{CNN-BiLSTM}
    \includegraphics[width=\textwidth]{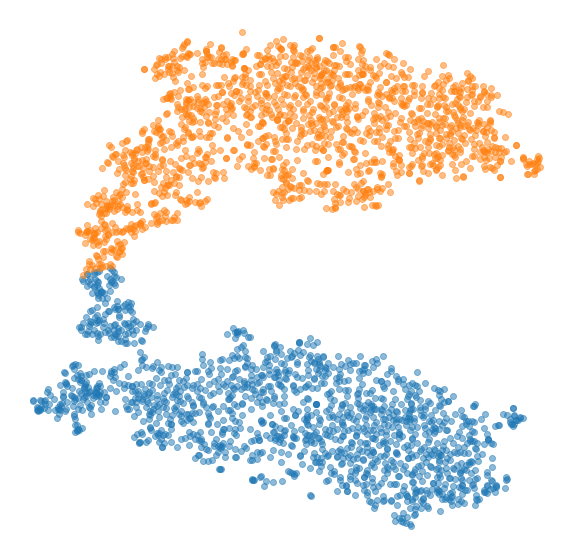}
    \label{fig:7-train-saroco}
\end{subfigure}

\begin{subfigure}[b]{0.1375\textwidth}
    \centering
    \includegraphics[width=\textwidth]{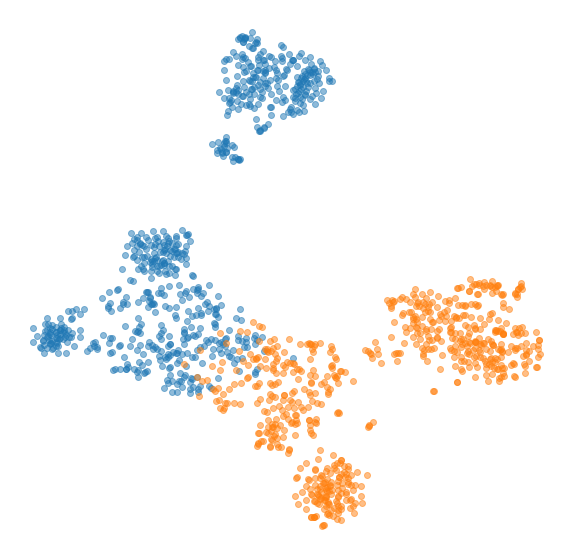}
    \label{fig:1-train-laroseda}
\end{subfigure}
\begin{subfigure}[b]{0.1375\textwidth}
    \centering
    \includegraphics[width=\textwidth]{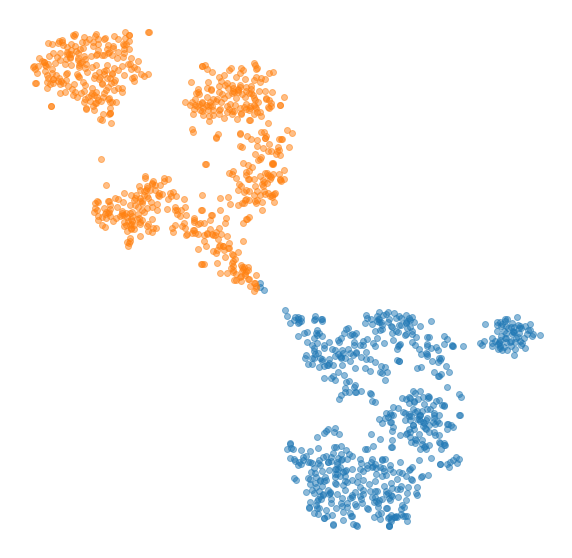}
    \label{fig:2-train-laroseda}
\end{subfigure}
\begin{subfigure}[b]{0.1375\textwidth}
    \centering
    \includegraphics[width=\textwidth]
    {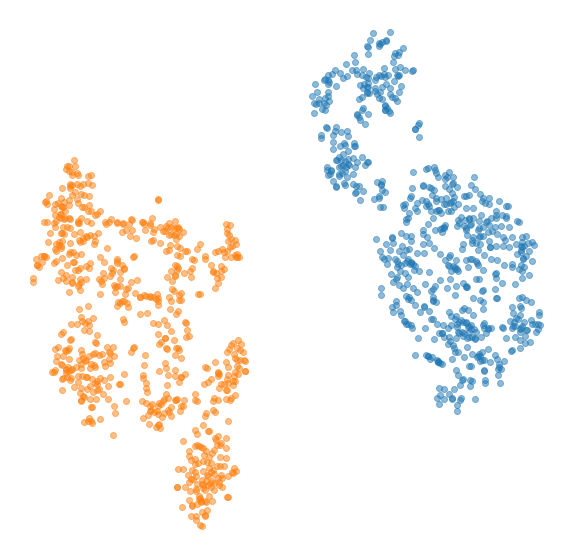}
    \label{fig:3-train-laroseda}
\end{subfigure}
\begin{subfigure}[b]{0.1375\textwidth}
    \centering
    \includegraphics[width=\textwidth]{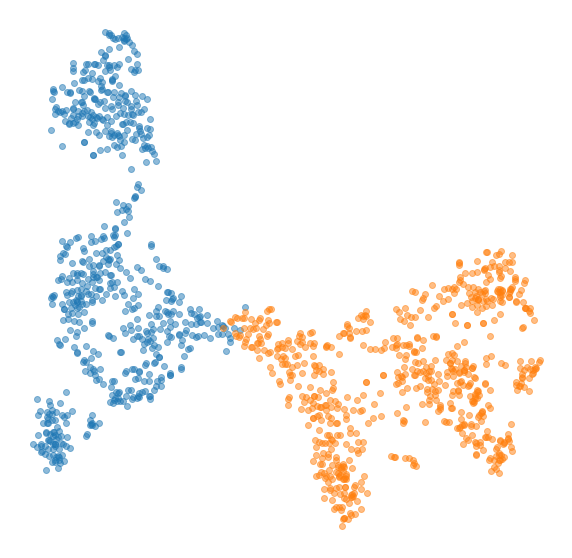}
    \label{fig:4-train-laroseda}
\end{subfigure}
\begin{subfigure}[b]{0.1375\textwidth}
    \centering
    \includegraphics[width=\textwidth]{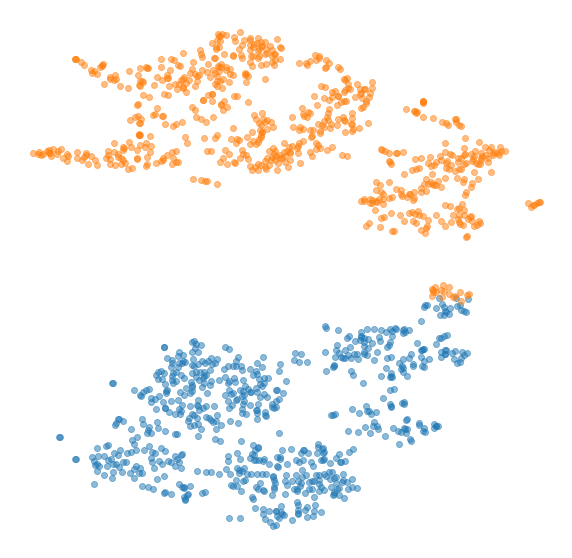}
    \label{fig:5-train-laroseda}
\end{subfigure}
\begin{subfigure}[b]{0.1375\textwidth}
    \centering
    \includegraphics[width=\textwidth]{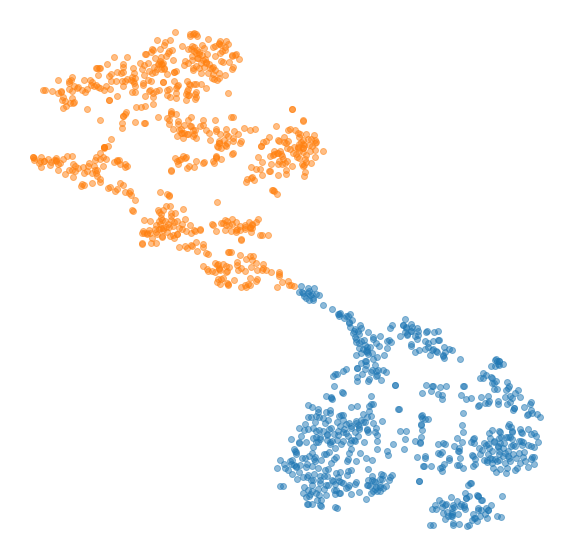}
    \label{fig:6-train-laroseda}
\end{subfigure}
\begin{subfigure}[b]{0.1375\textwidth}
    \centering
    \includegraphics[width=\textwidth]{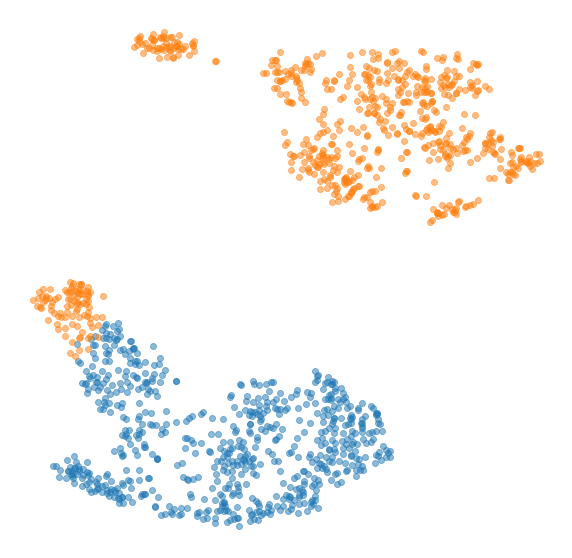}
    \label{fig:7-train-laroseda}
\end{subfigure}

\caption{t-SNE plots for each sub-module from the best-performing adversarial capsule network. The first row depicts the evaluation on SaRoCo, where blue indicates negative sentiment and orange represents positive one. The second row is for LaRoSeDa, where blue is for the non-satirical text, and orange is for the satirical one. The higher density on SaRoCo is because of a larger test dataset.}
\label{fig:tsne}
\end{figure}

We further assess the feature representation quality for each sub-module using the two-dimensional t-SNE visualisations upon the best-performing training results. Figure~\ref{fig:tsne} shows different clustering representations in most cases. For the SaRoCo dataset, the best delimitation is observed on the BiGRU sub-module, which is validated by the best performance achieved for the NLPL embeddings as shown in Table~\ref{tab:training_results}. A similar effect applies to the BiGRU sub-module trained and evaluated on LaRoSeDa.
Considering these results, the next set of experiments is performed based on the higher performance achieved with and without BERT embeddings, namely, the BiGRU sub-module with RoBERT and NLPL embeddings, respectively.

\begin{table}[!t]
\begin{center}
\caption{Accuracy (Acc) of the generic adversarial capsule network with different word embeddings and sub-modules.}
\label{tab:training_results}
\small
\begin{tabular}{clcc|cc}
\hline
\multirow{2}{*}{\textbf{Embeddings}} & \multicolumn{1}{|c|}{\multirow{2}{*}{\textbf{Sub-module}}} & \multicolumn{2}{c|}{\textbf{SaRoCo}} & \multicolumn{2}{c}{\textbf{LaRoSeDa}} \\ \cline{3-6} 
 & \multicolumn{1}{|c|}{} & \multicolumn{1}{c}{\textbf{Valid. Acc(\%)}} & \multicolumn{1}{c|}{\textbf{Test Acc(\%)}} & \multicolumn{1}{c}{\textbf{Valid. Acc(\%)}} & \multicolumn{1}{c}{\textbf{Test Acc(\%)}} \\ \hline
\multicolumn{1}{c|}{\multirow{7}{*}{\begin{tabular}[c]{@{}c@{}}CC\\      (300)\end{tabular}}} & \multicolumn{1}{l|}{CNN} & 95.57 & 95.34 & \textbf{95.52} & 95.19 \\
\multicolumn{1}{c|}{} & \multicolumn{1}{l|}{GRU} & 95.92 & 95.70 & 95.29 & 95.33 \\
\multicolumn{1}{c|}{} & \multicolumn{1}{l|}{BiGRU} & \textbf{96.02} & \textbf{95.80} & 95.19 & \textbf{95.53} \\
\multicolumn{1}{c|}{} & \multicolumn{1}{l|}{CNN-BiGRU} & 95.90 & 95.60 & 95.16 & 94.39 \\
\multicolumn{1}{c|}{} & \multicolumn{1}{l|}{LSTM} & 95.70 & 95.54 & 95.09 & 94.53 \\
\multicolumn{1}{c|}{} & \multicolumn{1}{l|}{BiLSTM} & 95.67 & 95.47 & 95.19 & 94.46 \\
\multicolumn{1}{c|}{} & \multicolumn{1}{l|}{CNN-BiLSTM} & 95.57 & 95.00 & 95.09 & 95.06 \\ \hline
\multicolumn{1}{c|}{\multirow{7}{*}{\begin{tabular}[c]{@{}c@{}}CoRoLa\\      (300)\end{tabular}}} & \multicolumn{1}{l|}{CNN} & 95.49 & 95.60 & 95.19 & 95.26 \\
\multicolumn{1}{c|}{} & \multicolumn{1}{l|}{GRU} & 95.97 & \textbf{95.77} & 95.39 & 95.59 \\
\multicolumn{1}{c|}{} & \multicolumn{1}{l|}{BiGRU} & \textbf{95.99} & \textbf{95.77} & 95.46 & \textbf{95.60} \\
\multicolumn{1}{c|}{} & \multicolumn{1}{l|}{CNN-BiGRU} & 95.82 & 95.67 & 95.42 & 95.19 \\
\multicolumn{1}{c|}{} & \multicolumn{1}{l|}{LSTM} & 95.85 & 95.70 & 95.39 & 94.86 \\
\multicolumn{1}{c|}{} & \multicolumn{1}{l|}{BiLSTM} & 95.90 & 95.70 & \textbf{95.56} & 95.26 \\

\multicolumn{1}{c|}{} & \multicolumn{1}{l|}{CNN-BiLSTM} & 95.65 & 95.50 & 95.52 & 94.73 \\ \hline
\multicolumn{1}{c|}{\multirow{7}{*}{\begin{tabular}[c]{@{}c@{}}NLPL\\      (100)\end{tabular}}} & \multicolumn{1}{l|}{CNN} & 95.79 & 95.80 & 95.29 & 95.86 \\
\multicolumn{1}{c|}{} & \multicolumn{1}{l|}{GRU} & 96.04 & 95.80 & \textbf{95.92} & \textbf{96.06} \\
\multicolumn{1}{c|}{} & \multicolumn{1}{l|}{BiGRU} & \textbf{96.10} & \textbf{96.15} & 95.79 & 95.83 \\
\multicolumn{1}{c|}{} & \multicolumn{1}{l|}{CNN-BiGRU} & 95.60 & 95.80 & 95.32 & 95.19 \\
\multicolumn{1}{c|}{} & \multicolumn{1}{l|}{LSTM} & 95.74 & 95.64 & 95.52 & 95.79 \\
\multicolumn{1}{c|}{} & \multicolumn{1}{l|}{BiLSTM} & 95.44 & 95.70 & 95.29 & 94.99 \\
\multicolumn{1}{c|}{} & \multicolumn{1}{l|}{CNN-BiLSTM} & 95.45 & 95.57 & 95.22 & 95.39 \\ \hline

\multicolumn{1}{c|}{\multirow{7}{*}{\begin{tabular}[c]{@{}c@{}}RoBERT\\      (768)\end{tabular}}} & \multicolumn{1}{l|}{CNN} & 98.17 & 98.09 & 98.50 & 98.56 \\
\multicolumn{1}{c|}{} & \multicolumn{1}{l|}{GRU} & 98.07 & 98.17 & 98.39 & 98.49 \\
\multicolumn{1}{c|}{} & \multicolumn{1}{l|}{BiGRU} & \textbf{98.27} & \textbf{98.32} & \textbf{98.54} & \textbf{98.60} \\
\multicolumn{1}{c|}{} & \multicolumn{1}{l|}{CNN-BiGRU} & 98.07 & 98.24 & 98.42 & 98.45 \\
\multicolumn{1}{c|}{} & \multicolumn{1}{l|}{LSTM} & 98.07 & 98.24 & 98.36 & 98.39 \\
\multicolumn{1}{c|}{} & \multicolumn{1}{l|}{BiLSTM} & 98.10 & 98.17 & 98.46 & 98.49 \\
\multicolumn{1}{c|}{} & \multicolumn{1}{l|}{CNN-BiLSTM} & 98.04 & 98.24 & 98.46 & 98.52 \\ \hline \hline
\multicolumn{2}{c|}{BERT-ro~\cite{rogoz-etal-2021-saroco}} & 82.41 & 73.00 & - & - \\
\multicolumn{2}{c|}{Char-CNN~\cite{rogoz-etal-2021-saroco}} & 73.42 & 69.66 & - & - \\ \hline \hline
\multicolumn{2}{c|}{HISK+BOWE-BERT+SOMs~\cite{tache-etal-2021-loresda}} & - & - & - & 90.90 \\ \hline
\end{tabular}
\end{center}
\end{table}

\textbf{Comparison to Existing Methods.}
The results of Rogoz et al.~\cite{rogoz-etal-2021-saroco} on the SaRoCo dataset show a more than 25\% gain for our models compared to the BERT-ro approach, while our models outperform the character-level CNN by more than 29\%. Human performance is a notable figure in deciding whether a selection of 200 news articles extracted from the dataset is satirical. Rogoz et al.~\cite{rogoz-etal-2021-saroco} explored the idea, involving ten human annotators and indicated that the human performance is at 87.35\% accuracy. Our approach surpasses this result by more than 11\%. In addition, the results shown by Tache et al.~\cite{tache-etal-2021-loresda} on the LaRoSeDa dataset prove the competitive performance of our proposed approach. Thus, our results are 7-8\% higher than their best model, HISK+BOWE-BERT+SOMs, which comprises histogram intersection string kernels, bag-of-words with BERT embeddings, and self-organizing maps.

\begin{table}[!t]
\begin{center}
\caption{Accuracy for various capsule hyperparameters.}
\label{tab:param_variation_results}
\small
\begin{tabular}{c|lll|lll}
\hline
\multirow{2}{*}{\textbf{Dataset}} & \multicolumn{3}{c|}{\textbf{N\textsubscript{pc}}} & \multicolumn{3}{c}{\textbf{N\textsubscript{cc}}} \\ \cline{2-7} 
 & \multicolumn{1}{c}{\textbf{2}} & \multicolumn{1}{c}{\textbf{8}} & \multicolumn{1}{c|}{\textbf{32}} & \multicolumn{1}{c}{\textbf{32}} & \multicolumn{1}{c}{\textbf{128}} & \multicolumn{1}{c}{\textbf{256}} \\ \hline
SaRoCo & 96.07 & 96.13 & \textbf{96.17} & 95.95 & \textbf{96.02} & 96.00 \\ \hline 
LaRoSeDa & 95.23 & \textbf{95.52} & 95.50 & 95.01 & \textbf{95.46} & 95.12 \\ \hline 
\end{tabular}
\end{center}
\end{table}

\textbf{Capsule Hyperparameter Variation.}
Figure~\ref{fig:caps-archi} depicts the hyperparameters of the capsule layers of our generic network, represented by $N_{pc}$ (i.e., the number of primary capsules) and $N_{cc}$ (i.e., the number of condensed capsules). We test the impact of these hyperparameters on the BiGRU sub-module with NLPL embeddings. We present the average for three runs per experiment. 
The chosen values for the  hyperparameters are $N_{pc}=\{2, 8, 32\}$ and $N_{cc}=\{32, 128, 256\}$ (see Table~\ref{tab:param_variation_results}).

During experiments, we observed that large values for $N_{pc}$ considerably impact the training time. This is mainly due to the operations over high-dimensional matrices in the $squash(\cdot)$ function from the iterative Dynamic Routing algorithm (see Equation~\ref{eq:3.5}).
Results from Table~\ref{tab:param_variation_results} support the intuition that a larger $N_{pc}$ would bring better results. The model trained on SaRoCo with $N_{pc}=32$ achieves the highest accuracy of 96.17\%; nevertheless, the difference between choosing 8 and 32 is minimal. For SaRoCo and LaRoSeDa, the best overall performance is achieved in a setting with $N_{cc}=128$, attaining accuracy scores of 96.02\% and 95.46\%, respectively. Based on both sets of results, we note that, for better performance, a hyperparameter search should be extended to the capsule hyperparameters.

\textbf{Ablation Study.}
Motivated by the noteworthy closeness in performance between the BiGRU-based models with NLPL and RoBERT embeddings, respectively, we perform an ablation study, slicing the generic model into four categories: baselines (i.e., NLPL-BiGRU and RoBERT-BiGRU), adversarial (Adv), Capsule, and Adv+Capsule. 
The best results on the test datasets are brought by the most complex models in terms of training and architecture, with a 96.02\% test accuracy for SaRoCo and a 95.82\% test accuracy for LaRoSeDa using the NLPL embeddings, as well as a 98.30\% test accuracy for SaRoCo and a 98.61\% test accuracy for LaRoSeDa using the RoBERT embeddings (see Table~\ref{tab:ablation_results}).

Regarding model complexity, we determine that except for the adversarial training on a baseline BiGRU model, the performance improves when capsule layers are added on top of it, irrespective of including the perturbed data in training. The increase in performance on the SaRoCo dataset  with our model is by 0.45\% for the NLPL embeddings and by 0.10\% for the RoBERT embeddings. We observe a decrease of 2.73\% when the most undersized model (i.e., NLPL-BiGRU) is compared with the most complex one (i.e., RoBERT-BiGRU+Adv+Capsule). For the LaRoSeDa dataset, we gain 1.18\% using the NLPL embeddings and 0.45\% with the RoBERT embeddings, respectively. Also, the test accuracy difference between the most complex and the most undersized models is 3.97\%, determining that the network conveys more value for the sentiment analysis task.

\begin{figure}[!t]
\centering
\newcommand{\figsizetsne}{0.155\textwidth} 
\newcommand{\tabhsize}{\hspace{0.5em}} 
\newcommand{\tabvsize}{2em} 
\newcommand{\manualrow}{5} 
\small
\begin{tabular}{c >{\tabhsize} c >{\tabhsize} c >{\tabhsize}c >{\tabhsize}c >{\tabhsize}c}
    & & Baseline & Adv & Capsule & Adv + Capsule \\

    \multirow{\manualrow}{*}{\rotatebox[origin=c]{90}{SaRoCo}} &
    \rotatebox[origin=c]{90}{NLPL} & 
    \includegraphics[width=\figsizetsne,valign=m]{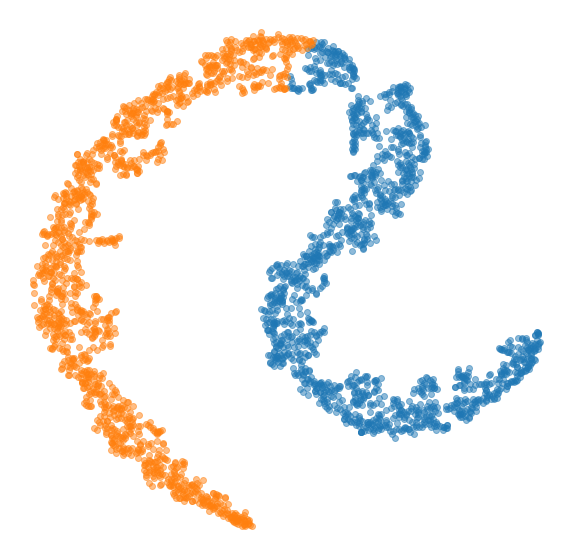} &
    \includegraphics[width=\figsizetsne,valign=m]{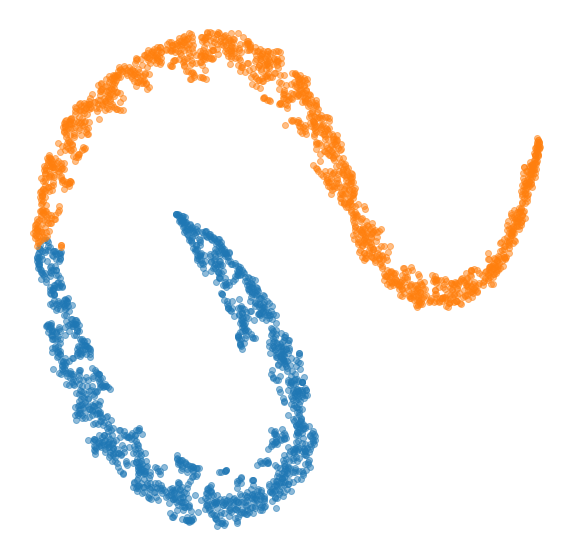} &
    \includegraphics[width=\figsizetsne,valign=m]{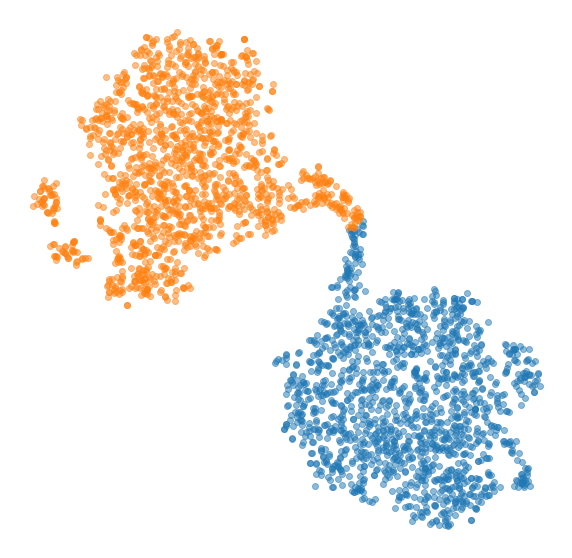} &
    \includegraphics[width=\figsizetsne,valign=m]{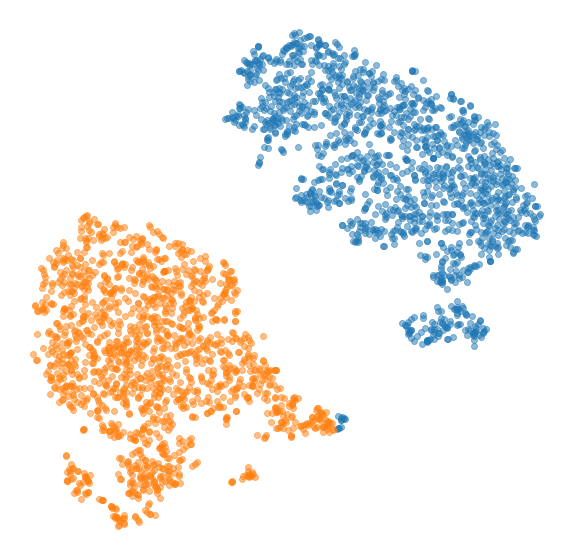} \\ [\tabvsize]

    &
    \rotatebox[origin=c]{90}{RoBERT} & 
    \includegraphics[width=\figsizetsne,valign=m]{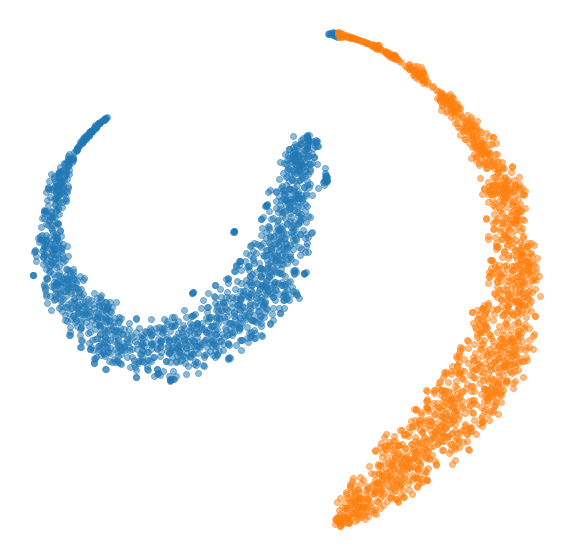} &
    \includegraphics[width=\figsizetsne,valign=m]{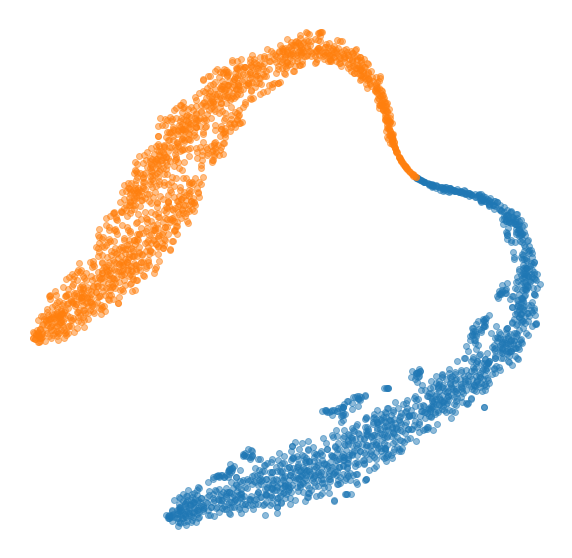} &
    \includegraphics[width=\figsizetsne,valign=m]{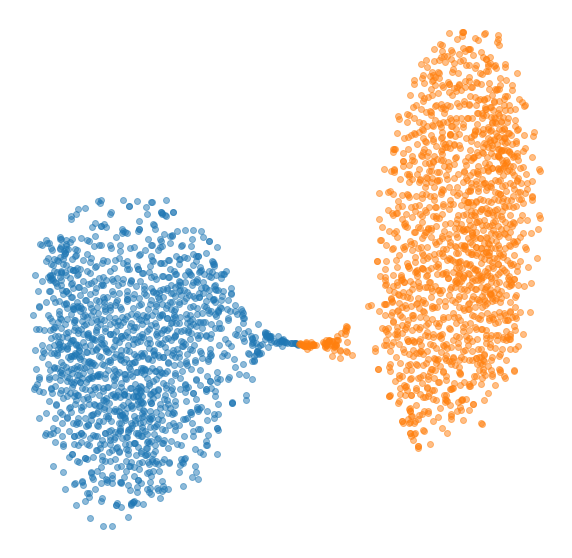} &
    \includegraphics[width=\figsizetsne,valign=m]{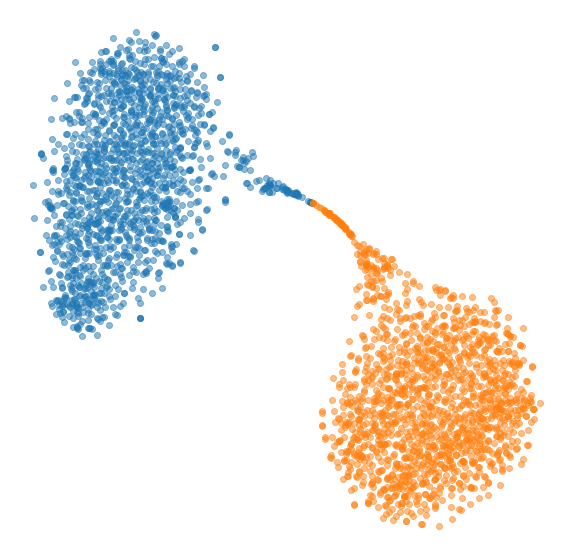} \\ [\tabvsize]

    \multirow{\manualrow}{*}{\rotatebox[origin=c]{90}{LaRoSeDa}} &
    \rotatebox[origin=c]{90}{NLPL} & 
    \includegraphics[width=\figsizetsne,valign=m]{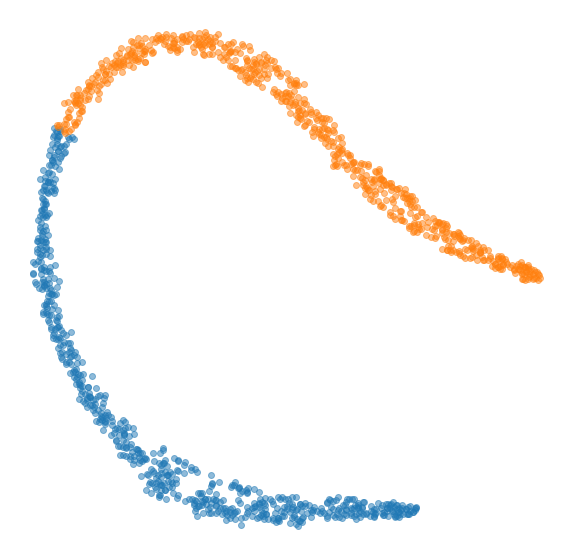} &
    \includegraphics[width=\figsizetsne,valign=m]{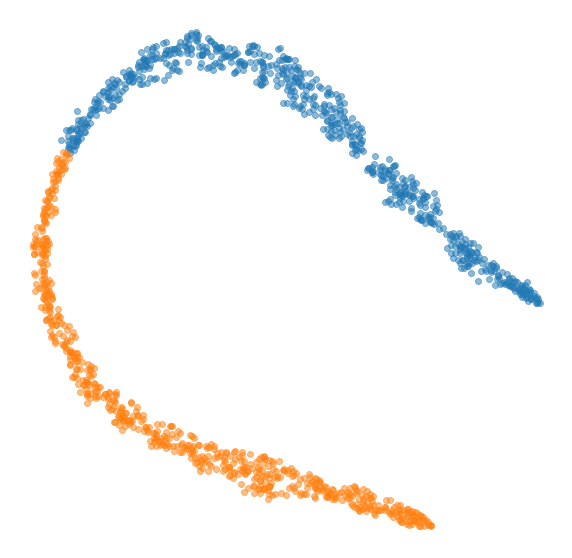} &
    \includegraphics[width=\figsizetsne,valign=m]{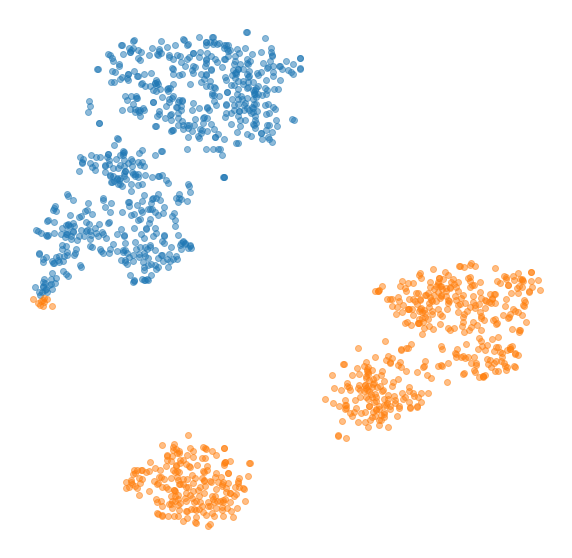} &
    \includegraphics[width=\figsizetsne,valign=m]{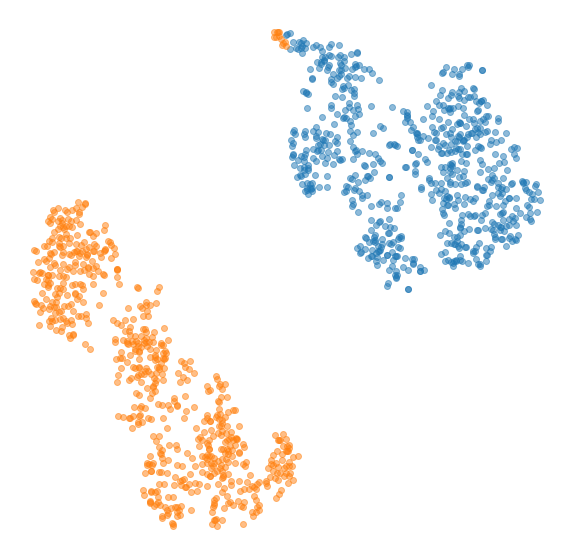} \\ [\tabvsize]

    &
    \rotatebox[origin=c]{90}{RoBERT} & 
    \includegraphics[width=\figsizetsne,valign=m]{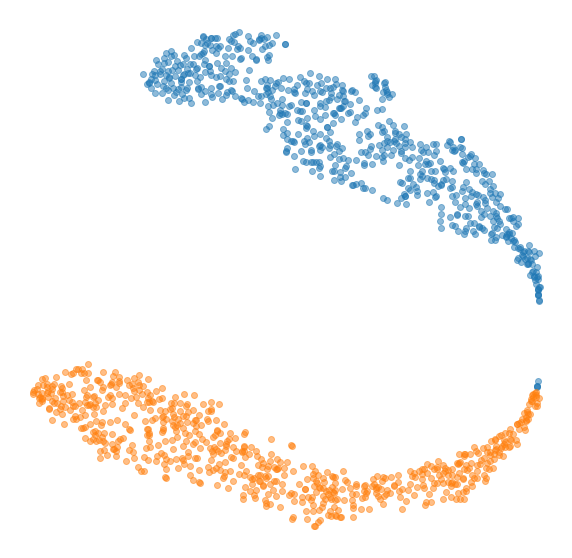} &
    \includegraphics[width=\figsizetsne,valign=m]{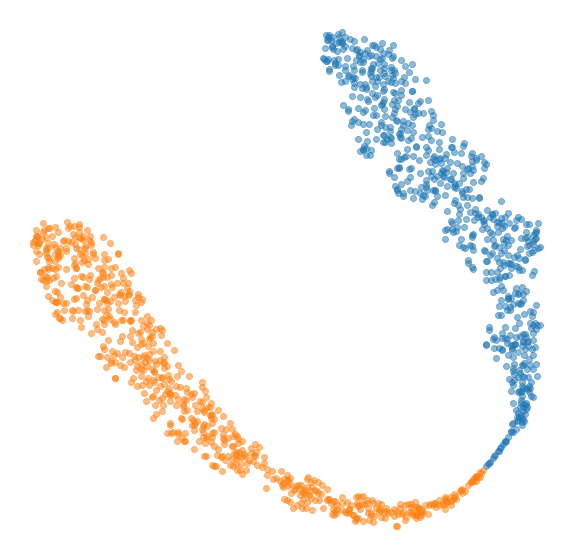} &
    \includegraphics[width=\figsizetsne,valign=m]{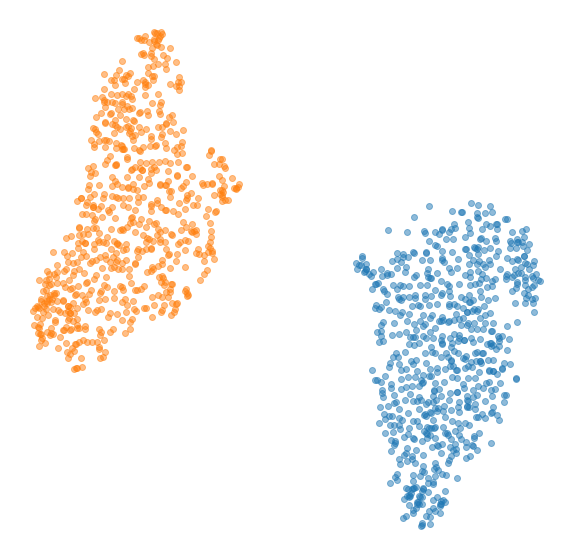} &
    \includegraphics[width=\figsizetsne,valign=m]{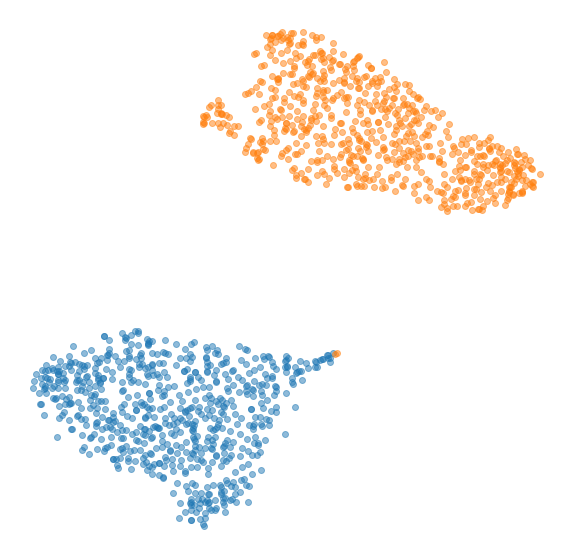} \\
\end{tabular}

\caption{t-SNE plots on embedding space for each model from the ablation study.}
\label{fig:tsne-ablation}
\end{figure}

The two-dimensional t-SNE embeddings depicted in Figure~\ref{fig:tsne-ablation} show the contrast between the capsule- and non-capsule-based models. The embeddings obtained with the BiGRU alone feature a specific chained distribution, with clusters defined by halving the sequence. The RoBERT embeddings convey a similar partition. In contrast, the capsule networks will mostly feature well-separated embedding clusters. No significant embedding change occurs when adversarial training is included.

\begin{table}[!t]
\begin{center}
\caption{Ablation study.}
\label{tab:ablation_results}
\small
\begin{tabular}{l|cc|cc}
\hline
\multicolumn{1}{c|}{\multirow{2}{*}{\textbf{Model}}} & \multicolumn{2}{c|}{\textbf{SaRoCo}} & \multicolumn{2}{c}{\textbf{LaRoSeDa}} \\ \cline{2-5} 
\multicolumn{1}{c|}{} & \textbf{Valid. Acc(\%)} & \textbf{Test Acc(\%)} & \textbf{Valid. Acc(\%)} & \textbf{Test Acc(\%)} \\ \hline
\multicolumn{1}{l|}{NLPL-BiGRU} & 94.80 & 95.57 & 92.73 & 94.64 \\
\multicolumn{1}{l|}{\tabspace +Adv} & 95.17 & 95.50 & 93.17 & 95.30 \\
\multicolumn{1}{l|}{\tabspace +Capsule} & 95.57 & 95.67 & 93.61 & 95.67 \\
\multicolumn{1}{l|}{\tabspace +Adv+Capsule} & \textbf{95.90} & \textbf{96.02} & \textbf{95.61} & \textbf{95.82} \\ \hline
\multicolumn{1}{l|}{RoBERT-BiGRU} & 98.23 & 98.20 & 98.68 & 98.16 \\ 
\multicolumn{1}{l|}{\tabspace +Adv} & \textbf{98.47} & 98.00 & \textbf{98.83} & 97.94 \\
\multicolumn{1}{l|}{\tabspace +Capsule} & 98.33 & 98.27 & 98.68 & 98.46 \\
\multicolumn{1}{l|}{\tabspace +Adv+Capsule} & 98.45 & \textbf{98.30} & 98.75 & \textbf{98.61} \\ \hline
\end{tabular}
\end{center}
\end{table}

\begin{table}[!t]
\begin{center}
\caption{Results for RoBERT-BiGRU augmented with RoGPT-2 data in terms of precision (P), recall (R), and accuracy (Acc).}
\label{tab:data-augmentation}
\small
\begin{tabular}{c|c|c|ccc}
\hline
\multicolumn{1}{l|}{\textbf{Dataset}} & \multicolumn{1}{l|}{\textbf{Decoder Method}} & \textbf{No. of Aug.} & \textbf{P(\%)} & \textbf{R(\%)} & \textbf{Acc(\%)} \\ \hline
\multirow{8}{*}{SaRoCo} & \multirow{4}{*}{Greedy} & 1,000 & 98.15 & 98.18 & 98.16 \\
 &  & 2,500 & 98.21 & 98.09 & 98.15 \\
 &  & 5,000 & 98.36 & 98.20 & 98.31 \\
 &  & 10,000 & 99.06 & 99.08 & \textbf{99.08} \\ \cline{2-6} 
 & \multirow{4}{*}{Beam-search-2} & 1,000 & 98.24 & 98.08 & 98.23 \\
 &  & 2,500 & 98.37 & 98.29 & 98.34 \\
 &  & 5,000 & 98.19 & 98.08 & 98.17 \\
 &  & 10,000 & 98.58 & 98.65 & \textbf{98.68} \\ \hline 
\multirow{8}{*}{LaRoSeDa} & \multirow{4}{*}{Greedy} & 1,000 & 98.39 & 98.31 & 98.36 \\
 &  & 2,500 & 98.82 & 98.52 & 98.70 \\
 &  & 5,000 & 98.85 & 98.77 & 98.87 \\
 &  & 10,000 & 98.94 & 98.87 & \textbf{98.94} \\ \cline{2-6} 
 & \multirow{4}{*}{Beam-search-2} & 1,000 & 98.44 & 98.40 & 98.43 \\
 &  & 2,500 & 98.72 & 98.49 & 98.64 \\
 &  & 5,000 & 98.90 & 98.80 & \textbf{98.87} \\
 &  & 10,000 & 98.82 & 98.70 & 98.77 \\ \hline
\end{tabular}
\end{center}
\end{table}

\begin{table}[!]
\centering
\caption{Examples from LaRoSeDa predicted with RoBERT-BiGRU. Ground truth (GT), Predicted (Pred) and Human labels are shown. P stands for Positive, N for Negative, and I for Indecisive.}
\label{tab:classification-analysis}
\begin{tabular}{p{1.82in}|p{1.82in}|ll|l}
\hline
\multicolumn{1}{c|}{\textbf{Romanian text}} & \multicolumn{1}{c|}{\textbf{English translation}} & \multicolumn{1}{c|}{\textbf{GT
}} & \multicolumn{1}{c|}{\textbf{Pred}} & \multicolumn{1}{c}{\textbf{Human}} \\ \hline
o boxa ok din punct de vedere calitate pret daca este cumparata de unde trebuie. aici nu apare nici numele complet al boxei iar descrierea este saraca, plus pretul cu mult peste cat o gasesti in alte magazine. & a good speaker in terms of quality and price if it is bought from the right place. the speaker's full name does not appear here and the description is poor, plus the price is much higher than what you can find in other stores. & \multicolumn{1}{l|}{P} & N & I \\ \hline
bun doar pentru incarcare (nu face conexiune, nu incarca rapid modelul nexus x). nu pare sa fie universal. nu realizeaza conexiune. voi mai incerca cu diverse cabluri micro usb si revin daca reusesc sa conectez telefonul la calculator. & good only for charging (doesn't connect, doesn't fast charge the nexus x model). it doesn't seem to be universal. it doesn't connect. I will try with various micro usb cables and return if I can connect the phone to the computer. & \multicolumn{1}{l|}{P} & N & N \\ \hline
imi place. o bratara feminina care isi face bine treaba. se sincronizeaza foarte bine cu android - samsung. bateria are autonomie zile cu functia pulse ox activata, fara aceasta functie scrie ca ar avea zile, dar nu am incercat. & I like it. a feminine bracelet that does its job well. it synchronizes very well with android - samsung. the battery has an autonomy of days with the pulse ox function activated, without this function, it says it would have days, but I have not tried it. & \multicolumn{1}{l|}{N} & P & P \\ \hline
aproape multumit. am cumparat acest produs in urma cu o luna si pana acum doua zile am fost foarte multumit de el. bateria asigura o autonomie de - zile, finisajele sunt ok. & almost satisfied. I bought this product a month ago and until two days ago I was very satisfied with it. the battery ensures the autonomy of - days, the finishes are ok. & \multicolumn{1}{l|}{N} & P & P \\ \hline
bun. folie calitativ buna dar nepotrivita pentru ecrane curbate. raman - milimetri dezlipiti pe margine. personal as recomanda folie de plastic pentru ecrane curbate dupa experienta asta. & good. good quality foil but not   suitable for curved screens. it remains - millimetres unglued on the edge. I would personally recommend a plastic film for curved screens after this experience. & \multicolumn{1}{l|}{N} & P & I \\ \hline
multumita! este foarte buna sunet clar! doar ca are probleme la conectarea cu bluetooth, il gaseste greu sau face nazuri km a conectare trebuie sa caut de multi ori bluetooth-ul. in rest e ok. & pleased! it is a very good clear sound! it's just that it has problems connecting with bluetooth, it finds it hard or it's difficult to connect, I have to look for bluetooth many times. the rest is ok. & \multicolumn{1}{l|}{N} & P & P \\ \hline
recomand. claritate, sunet bun si un microfon super, fara fire, doar o cutiuta miniona de incarcare! pretul este mult sub cel de la apple. multumit de produs. & I recommend it. clarity, good sound and a great microphone, no wires, just a tiny charging box! the price is much lower than that of apple. happy about the product. & \multicolumn{1}{l|}{N} & P & P \\ \hline
decent. il folosesc cu un samsung si nici pe departe nu are incarcare fast charge. daca nu te grabesti si ai rabdare sa astepti, merge. & decent. I use it with a Samsung, which doesn't even have a fast charge. It will work if you are not in a hurry and have the patience to wait. & \multicolumn{1}{l|}{N} & P & I \\ \hline
\end{tabular}
\end{table}

\textbf{Data Augmentation.}
Next, we incorporate the RoGPT-2 text continuation examples on a set of samples using two strategies for the decoder (i.e., greedy and beam-search-2). We perform experiments with the RoBERT-BiGRU model and show that the generative effort increases the overall performance for both tasks (see Table~\ref{tab:data-augmentation}). In most cases, the RoBERT embeddings bring increased performance on the LaRoSeDa dataset as a consequence of the polarized effect of the product reviews, being strongly positive or negative. This polarization impact also applies to the models trained on augmented data. 
Data augmentation using the greedy decoder method achieves the best performance on SaRoCo, with a 99.08\% test accuracy, employing 10,000 expanded texts, compared with the best accuracy of 98.68\% obtained with beam-search-2. Furthermore, on LaRoSeDa, we determine similar performance on the greedy search algorithm with the best accuracy of 98.94\% for 10,000 augmented texts. However, for the second dataset, more generated data will not necessarily determine the best performance as in the beam-search-2 scenario, using 10,000 augmented texts slightly underperforms in contrast with 5,000 examples. 

\textbf{Discussions.}
RoBERT-BiGRU, augmented with RoGPT-2 samples, correctly classifies 1,344 out of 1,362 examples from the LaRoSeDa test dataset. Due to spatial constraints, Table~\ref{tab:classification-analysis} depicts only the shortest eight misclassified texts out of 18, for which ground truth, predicted, and human annotated labels are shown. Two human annotators concluded from these examples that three indecisions and five classifications contradict the expected ones. The uncertain results and the negative misclassifications are expected to have been 3-out-of-5 stars ratings, which were assumed negative when the dataset was created. 
Furthermore, we observe strongly positive texts such as ``I like it. A feminine bracelet that does its job well", ``I was very satisfied with it", ``happy about the product", ``I recommend it", and ``pleased! it is a very good clear sound!" have negative ground truth in the dataset. However, these are positive examples for the model and human annotators. Thus, we determine noise in the LaRoSeDa dataset, which is expected for datasets gathered from online sources, as the origin of the noise can be introduced by the page user or by automated data extractors.

\section{Conclusions} 
Satire detection and sentiment analysis are important NLP tasks for which literature provides an ample palette of models and applications.
Despite the more polarization expected on the product review task in contrast with the increased passivity of satirical texts, our models properly encapsulate the meaning represented by relevant features.
In the syntactic and semantic context of our tasks, there is a slight difference in performance for the CC, CoRoLa, and NLPL embeddings, whereas fine-tuning the pre-trained RoBERT model brings up to 3\% performance improvement. 
We showed in many experiments that our parameterized capsule framework can be adapted to specific problems. Moreover, we can improve the capsule network by employing data augmentation using generative models such as RoGPT-2, achieving a maximum gain of 0.6\%. Based on our results, the potential of such an architecture is of increased significance, thus enabling further work in this direction.

\section*{Acknowledgements}
The research has been funded by the University Politehnica of Bucharest through the PubArt program. 

\bibliographystyle{splncs04}
\bibliography{bibliography}

\end{document}